\documentclass[conference]{IEEEtran}
\IEEEoverridecommandlockouts
\usepackage{cite}
\usepackage{amsmath,amssymb,amsfonts}
\usepackage{algorithmic}
\usepackage{graphicx}
\usepackage{textcomp}
\usepackage{xcolor}
\usepackage{xspace} 
\usepackage[]{algorithm}
\usepackage{multirow}
\usepackage{booktabs} 
\usepackage{arydshln}
\usepackage{relsize}
\usepackage{url}
\usepackage{lipsum}

\DeclareMathOperator*{\argmin}{argmin} 

\def\BibTeX{{\rm B\kern-.05em{\sc i\kern-.025em b}\kern-.08em
T\kern-.1667em\lower.7ex\hbox{E}\kern-.125emX}}
\begin{document}

\newcommand{\toolname}{ALWANN\xspace} 

\title{
 \vspace{-1.5em}\small To appear in 2019 IEEE/ACM International Conference on Computer-Aided Design (ICCAD).  \textcopyright 2019 IEEE \\\vspace*{1.2em}
\huge ALWANN: \underline{A}utomatic \underline{L}ayer-\underline{W}ise \underline{A}pproximation of \\Deep \underline{N}eural \underline{N}etwork Accelerators without Retraining
\vspace*{-0.5em}
}

\author{
    
    \IEEEauthorblockN{Vojtech Mrazek\IEEEauthorrefmark{1}\IEEEauthorrefmark{2}, Zdenek Vasicek\IEEEauthorrefmark{2}, Lukas Sekanina\IEEEauthorrefmark{2}, Muhammad Abdullah Hanif\IEEEauthorrefmark{1}, Muhammad Shafique\IEEEauthorrefmark{1}}
    \IEEEauthorblockA{\{mrazek, vasicek, sekanina\}@fit.vutbr.cz, \{muhammad.hanif, muhammad.shafique\}@tuwien.ac.at}
    \IEEEauthorblockA{\IEEEauthorrefmark{2}Faculty of Information Technology, IT4Innovations Centre of Excellence, Brno University of Technology, Czech Republic}
    \IEEEauthorblockA{\IEEEauthorrefmark{1} Technische Universit\"at Wien (TU Wien), Vienna, Austria}
    \vspace*{-3em}
}

\maketitle

\begin{abstract}
The state-of-the-art approaches employ approximate computing to reduce the energy consumption of DNN hardware. Approximate DNNs then require extensive retraining afterwards to recover from the accuracy loss caused by the use of approximate operations. However, retraining of complex DNNs does not scale well. In this paper, we demonstrate that efficient approximations can be introduced into the computational path of DNN accelerators while retraining can completely be avoided.
\toolname provides highly optimized implementations of DNNs for custom low-power accelerators in which the number of computing units is lower than the number of DNN layers. First, a fully trained DNN (e.g., in TensorFlow) is converted to operate with 8-bit weights and 8-bit multipliers in convolutional layers. A suitable approximate multiplier is then selected for each computing element from a library of approximate multipliers in such a way that (i) one approximate multiplier serves several layers, and (ii) the overall classification error and energy consumption are minimized. The optimizations including the multiplier selection problem are solved by means of a multiobjective optimization NSGA-II algorithm. In order to completely avoid the computationally expensive retraining of DNN, which is usually employed to improve the classification accuracy, we propose a simple weight updating scheme that compensates the inaccuracy introduced by employing approximate multipliers. The proposed approach is evaluated for two architectures of DNN accelerators with approximate multipliers from the open-source "EvoApprox" library, while executing three versions of ResNet on CIFAR-10. We report that the proposed approach saves 30\% of energy needed for multiplication in convolutional layers of ResNet-50 while the accuracy is degraded by only 0.6\% (0.9\% for the ResNet-14). 
The proposed technique and approximate layers are available as an open-source extension of TensorFlow at \url{https://github.com/ehw-fit/tf-approximate}.


\end{abstract}


\vspace{-1em}
\section{Introduction}
\vspace{-.2em}

Providing an energy efficient implementation of the inference engine of DNNs becomes
crucial to enable efficient data processing on smart devices or IoT edge nodes where the energy resources are typically limited.
The state-of-the-art works~\cite{chippa,hanif:jople2018} clearly indicate that neural networks feature an intrinsic error-resilience property which can be exploited by adopting the principles of 
\textit{approximate computing} to develop
energy-efficient hardware accelerators~\cite{mrazek:dac2019} of DNNs. 


\textbf{Research Questions:} 
In this work, we explore whether it is possible to achieve energy savings in the computational path (CP) of DNN hardware accelerators by means of introducing approximate arithmetic operators, \textbf{but without performing any time-exhaustive retraining.} The retraining is also impossible for proprietary NNs where the training set may not be available. We also ask whether there is a computationally inexpensive way to adapt the weights of an already trained DNN to its particular approximate implementation. {Another question is whether we have to apply the approximations uniformly across all the layers or search for a suitable approximation for each layer separately.} 

\begin{figure}[ht]
    \centering\vspace{-1em}
    \includegraphics[width=0.85\columnwidth]{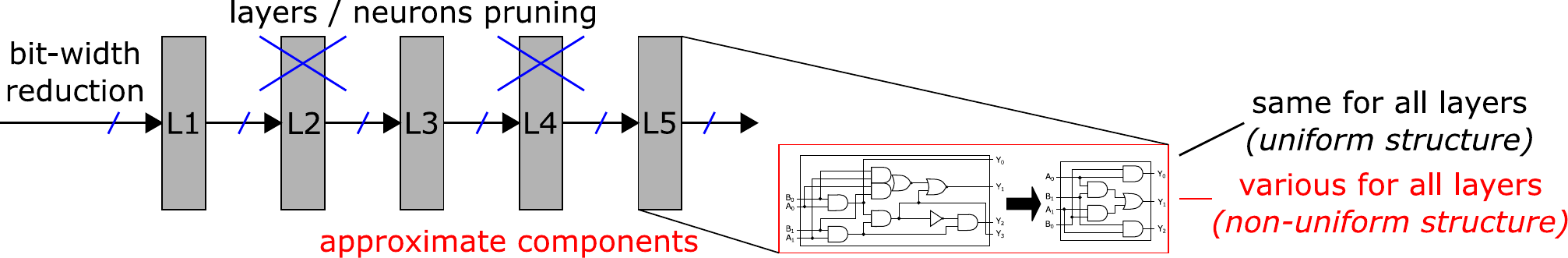}\vspace{-0.8em}
    \caption{Illustration of typical approximations of computational path of DNNs. The methods targeted in this work are marked red.}\vspace{-1em}
    \label{fig:approx}
\end{figure}

\textbf{Approximations of DNNs:}
The energy optimization of CP in DNN accelerators by means of approximations can be introduced on various levels~\cite{hanif:jople2018} (see Fig.~\ref{fig:approx}). 
The state-of-the-art \textit{Tensor Processor Unit (TPU)} accelerator accomplishes a satisfactory accuracy with 8-bit integer operations. If the precision is fixed, different approaches can be applied to make the computing more effective, e.g.,  pruning~\cite{Venkataramani:axnn,zhang:2015}, which involves removing some connections from the DNN or introducing approximate components into the CP~\cite{mrazek:iccad16,Sarwar:2018}.

In the prior research, significant energy savings resulting from introducing various approximations in the computational path were documented. For the multi-layer perceptron (MLP) NNs, the analysis helps to determine the candidate neurons for the approximations or pruning. But it turns out that computational structures are \textit{non-uniform}. In contrast with that, the approximate convolutional NNs utilized \textit{a uniform structure} where all layers employ the same components. 
Considering the fact that the layers differ in the error resilience~\cite{hanif:2018}, the goal of the proposed methodology is to find the right approximation level for the convolutional layers separately. 
In order to eliminate the error introduced by the approximations, 
the retraining (based on a feedback from a training set) is typically used to adapt the NNs to these approximations. 
\textit{However, as the simulation of approximate hardware components does not usually scale well, the simulation of the entire DNN is thus slowed down 
in one order of magnitude for CPU and in three orders for GPU} (see Sect \ref{sec:results}). Regarding to the results presented in~\cite{mrazek:iccad16}, the retraining on an approximate NN with 50 layers employing arbitrary approximate components would take 89 days. This issue was partly addressed by Sarwar et al.~\cite{Sarwar:2018} by using a very limited set of approximate components (having limited efficiency) whose simulation scaled well. If one is going to use arbitrary approximate components in DNNs, \textit{avoiding of the retraining} is necessary and it becomes one of the major research challenges addressed by this paper.

\textbf{Hyperparameters Tuning:}
The second challenge is how to automatically and effectively \textit{select the right approximate implementation} for each relevant component (e.g., layer). The majority of modern DNN architectures was constructed manually by designers. With the increasing computational resources, automated methodologies for hyperparameters tuning based on the grid search, random search, Bayesian optimization, and genetic algorithms (GA) become popular. 
Xie and Yuille~\cite{xie:cnn} proposed a GA-based tool for optimization of basic building modules of a large network. The generated DNN structures often perform better than the standard manually designed ones. Another GA-based tool, \textit{CoDeepNEAT}~\cite{miikkulainen:2017}, which develops the \textit{NEAT} approach for evolving MLP structures, provides a different abstraction --- it connects layers instead of neurons and tune the layer parameters (e.g., activation functions, kernel sizes etc.). Each candidate solution (represented by a graph) is transformed to a DNN. 
The training of candidate DNNs is performed by means of standard learning algorithms.
The extended version~\cite{liang:2019} allows to optimize multiple objectives (overall accuracy and complexity of the resulting NNs).


\textbf{Novel Contributions:} In this paper, we aim at developing \textit{\toolname framework} that takes a trained (frozen) NN and a set of approximate multipliers as inputs and generates Pareto set of approximate neural networks (AxNNs) that trades off accuracy and the energy. 

\textit{The proposed methodology is inspired by the hyperparameter tuning algorithms and genetic algorithms.} It optimizes two main design criteria for DNNs --- the overall DNN accuracy and the energy consumed by the approximate layers. It satisfies all constrains induced by a particular DNN accelerator, e.g., at most $T$ approximate execution units can be used, or the units can be power-gated or pipelined.

To avoid the time-critical or unavailable retraining process, we proposed a fast \textit{weight tuning} algorithm that adapts the layer weights to the employed multipliers and allows for improving the accuracy of NN by 4\% in the average. Thanks to the relatively fast evaluation of candidate implementations, the proposed methodology allows to significantly extend the complexity of the NNs that can be approximated --- up to 120M multiplications (ResNet-50) in contrast with 200k multiplications in LeNet-6~\cite{mrazek:iccad16}. 


The proposed approach enables to approximate DNNs with the results comparable to the results obtained by other automated approximation methods, but for more complex NNs and without retraining. Moreover, the proposed approach is capable of constructing more energy-efficient NNs than the approach based on a layer removal followed by a training from scratch~\cite{he:resnet}.

An open-source library of approximate convolutional layers is provided at \textcolor{blue}{ \url{https://github.com/ehw-fit/tf-approximate}}. The library extends the widely used TensorFlow framework with approximate layers and an example of layer replacement is provided.

%
%
%

\section{Background and related work}
\subsection{Approximate neural networks}
\begin{figure}[ht]
    \centering\vspace{-1em}
    \includegraphics[width=\columnwidth]{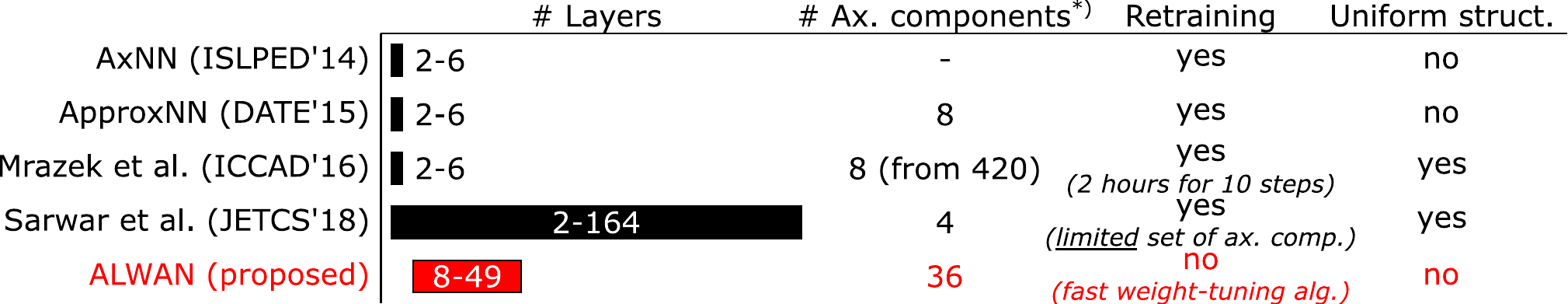}\vspace{-0.8em}
    \caption{Complexity  and properties of the recently publishedn    NN CP approximation methods compared to the proposed algorithm (see Sec. \ref{sec:exsetup} for details). \textsuperscript{*)} The number of approximate components is reported for 8-bit inputs.}\vspace{-1em}
    \label{fig:complex}
\end{figure}

A straightforward approach for the automated construction of NNs with approximate CP is to optimize the bit precision for the data structures used in NN \cite{gysel2018ristretto}; a recent research shows that in specific cases one bit can be sufficient to represent the weights~\cite{bnn:16}. 
Let us suppose that the bit width is fixed to $n$ bits due to architectural constraints. There are several ways how to improve the energy efficiency of the $n$-bit arithmetic operations (Fig.~\ref{fig:complex}).

Ventkatamani et al.~\cite{Venkataramani:axnn} identified error-resilient neurons based on gradients calculated during the training process of a MLP. For these neurons, an approximation based on the precision modification and simplifying of activation functions was applied. 
Due to these modifications additional retraining of the approximate MLP was required.

Zhang et al.~\cite{zhang:2015} used a different approach for the critical neuron identification. A neuron is considered as critical, if small jitters on the neuron’s computation introduce a large output quality degradation; otherwise, the neuron is resilient. For the resilient neurons, the memory access skipping, precision scaling or arithmetic operation approximations are applied. To increase the overall accuracy, the resulting neural networks were retrained. This approach was only evaluated on a MLP.

In the case of CNNs, Mrazek et al.~\cite{mrazek:iccad16} introduced approximate multipliers to convolutional layers of the LeNet neural network. They showed  that the back-propagation algorithm can adapt the weights of CNN to the used approximate multipliers and significant power saving can be achieved for a negligible loss in accuracy. Approximate multipliers based on the principles of multiplierless multiplication were introduced to complex CNNs in~\cite{Sarwar:2018}.
The authors modified the learning algorithm in such a way that only those weights could be used for which an efficient implementation of approximate multiplication exists. 
The authors showed that the approximations can provide significant power savings in the computational path even for deep neural networks. However, the major limitation of this approach is that arbitrary approximate multiplier cannot be introduced to the NN.


Although the aforementioned approximation methods decrease the accuracy of the NNs, the resulting NNs can be beneficial for other approaches, for example, in progressive chain classifiers (PCCs)~\cite{Choi2019}. In PCCs, there is a chain of classifier models that progressively grow in complexity and
accuracy. 
After evaluating a stage it is checked whether its confidence is high; if so the remaining stages of the PCC are not evaluated.

\subsection{Neural network accelerators} \label{sec:systolic}


Numerous accelerator designs have been proposed for accelerating DNN inference \cite{chen2014dadiannao, jouppi2017datacenter, chen2017eyeriss, lu2017flexflow}. Almost all of these accelerators mainly focus on accelerating the dot product operation, which is the fundamental operation in the convolutional and fully-connected layers of the NNs, i.e., the most computationally and memory intensive layers in NNs~\cite{chen2014dadiannao}. 
Accelerators like TPU~\cite{jouppi2017datacenter} and Flexflow~\cite{lu2017flexflow} perform layer-wise computations while others like DaDianNao~\cite{chen2014dadiannao} focus on a pipelined implementation of a network for achieving significant efficiency gains as compared to CPUs and GPUs. 

\begin{figure}[ht]
    \centering\vspace{-1em}
    \includegraphics[width=0.9\columnwidth]{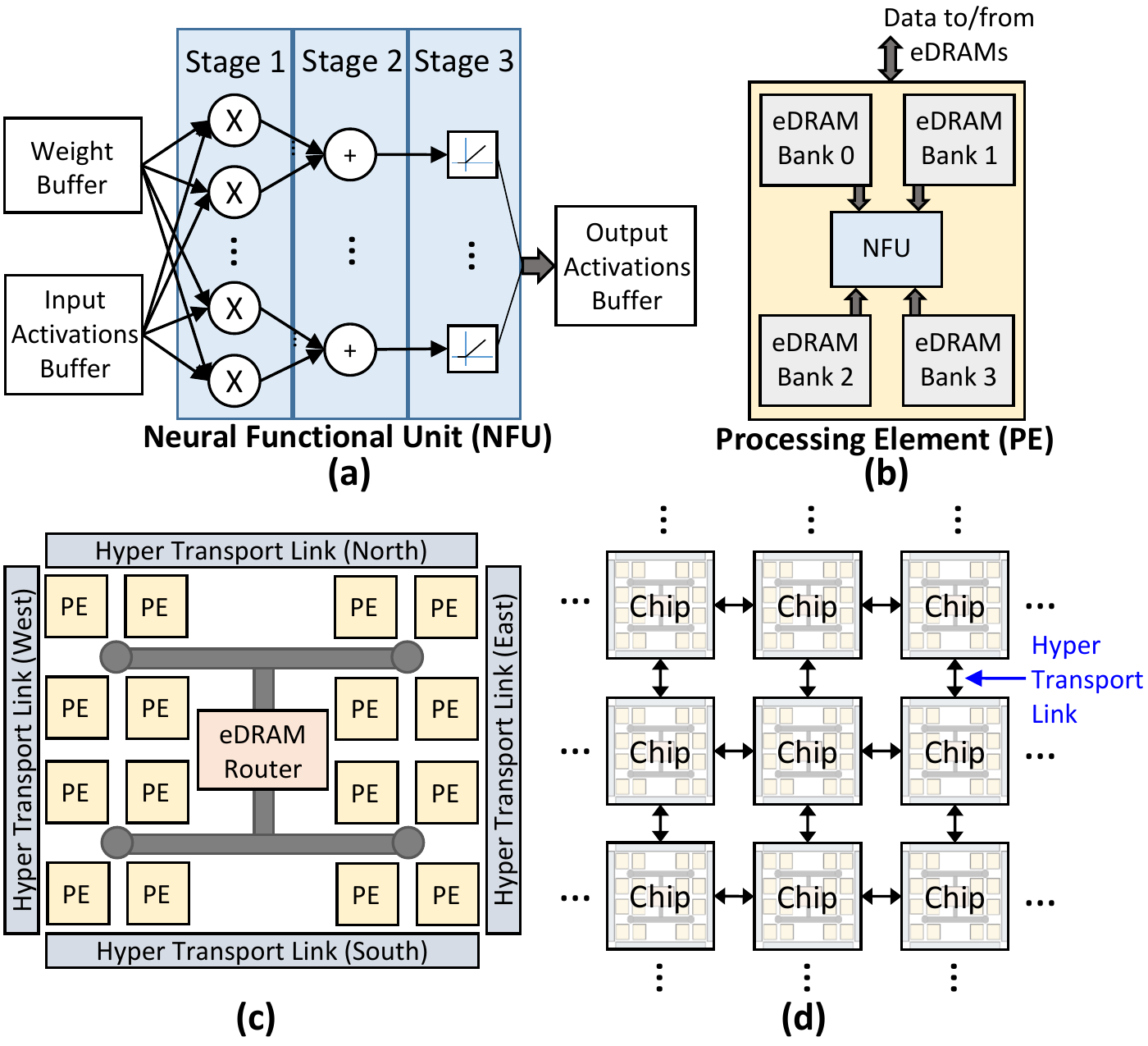}\vspace{-0.8em}
    \caption{DaDianNao accelerator design~\cite{chen2014dadiannao}: (a) Neural Functional Unit (NFU); (b) Processing Element; (c) Chip/Node; and (d) Interconnected chips.}\vspace{-1em}
    \label{fig:Acc_Fig_1}
\end{figure}

DaDianNao~\cite{chen2014dadiannao} is a promising architecture which makes use of distributed memory to reduce the high energy costs related to main memory accesses. The memory units are deployed near Neural Functional Units (NFUs), where the NFUs are pipelined version of the computations required in NN layers. An NFU has three main pipeline stages, i.e., (1) multiplication of weights with input activations; (2) additions of products; and (3) application of an activation function on the generated outputs. An illustration of the NFU is shown in Fig.~\ref{fig:Acc_Fig_1}a. An NFU combined with four banks of eDRAM forms a processing element (PE) (see Fig.~\ref{fig:Acc_Fig_1}b) where multiple PEs are connected together to form a node/chip (see Fig.~\ref{fig:Acc_Fig_1}c). The chips are then interconnected to form a system (see Fig.~\ref{fig:Acc_Fig_1}d) which is then used for deploying complete NNs. 

\begin{figure}[t]
    \centering\vspace{-0em}
    \includegraphics[width=0.9\columnwidth]{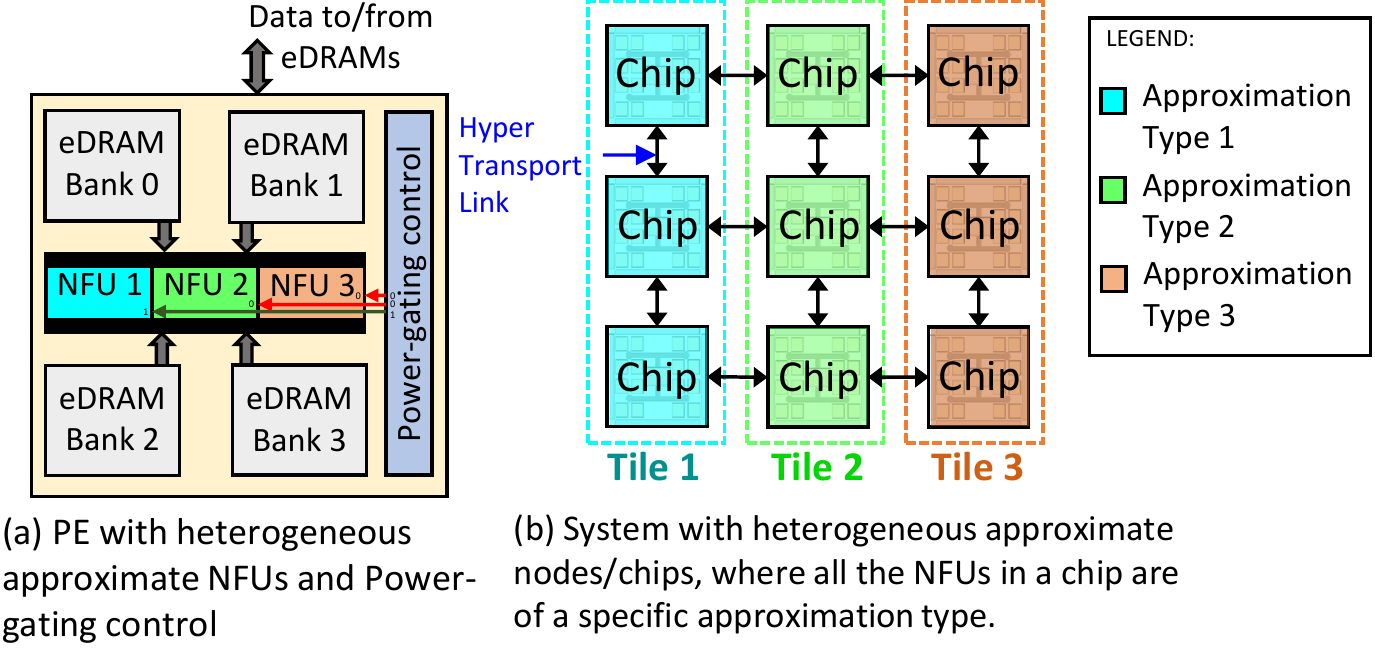}\vspace{-0.8em}
    \caption{(a) Scenario 1: All the chips are homogeneous, however, each PE has multiple approximate variants of NFUs installed in it out of which one is selected and rest are power-gated. (b) Scenario 2: Each PE contains a single approximate NFU and all the NFUs in a chip are of a specific approximations type, however, approximation type can vary across chips.}\vspace{-1.5em}
    \label{fig:Acc_Fig_2}
\end{figure}

From approximations point of view, the computational units in the NFUs can be approximated to improve the energy/power efficiency of the system. There can be multiple scenarios for employing approximations in the computational units: (1) a single chip contains more than one type of approximate computational units, however, at run-time, only one type is selected for operation while all the rest are power-gated; and (2) a single chip contains a specific type of approximate computational units. Both the highlighted scenarios are in line with the concept of heterogeneous approximate computing units proposed in~\cite{el2017embracing}. Exemplar hardware architectures for both the scenarios are illustrated in Fig.~\ref{fig:Acc_Fig_2}.  
Note that a layer of an NN can be mapped to one or more chips, which are jointly referred as \textit{tile} from henceforth. Moreover, in this work we assume that a layer can have only one type of approximation, therefore, all the chips in one tile are assumed to have the same configuration as highlighted in Fig.~\ref{fig:Acc_Fig_2}b.

Note that here we consider approximate computational units with fixed approximations, i.e., not configurable approximate modules like the one proposed in~\cite{mrazek:ahs2018}. This is due to the fact that the efficiency gains of the configurable modules are significantly affected by the required configurability characteristics.

\section{Proposed methodology}\label{sec:methodology}
\subsection{Overview}

\toolname requires the following inputs from the user: already trained NN being subject of the approximation, a library of basic approximate components (adders, multipliers, MACs) and knowledge of the architecture of the final HW accelerator. Two HW-based architectures (as discussed in the previous section) are considered in this work: \textit{pipelined} and \textit{power-gated} arrays.
For simplicity, the MAC units will be implemented using accurate addition and approximate multiplication, but approximate addition can be introduced as well in general.
Let $L=\{L_1,L_2,...\}$ be a set of indexes of convolutional layers of NN and $M$ be a set of available approximate $w$-bit multipliers. The user should specify the number of different tiles $|T|$ the accelerator will consists of. Typically, $|T| < |L|$ and $w=8$ is sufficient. Each tile's NFU consists of the array of the same MAC units. Each layer $L_i$ is supposed to be executed on a single tile $T_j$.

The method outputs a set of AxNNs (modified original NN together with the corresponding configuration of the HW accelerator tiles) that are Pareto optimal w.r.t. the energy consumption and classification accuracy. The approximations are introduced to the original NN by \textit{replacement} of the accurate convolutional layers by approximate ones together with \textit{weight tuning}. Considering the structure of the HW-based accelerator, two task are solved simultaneously. We are looking for the \textit{assignment of the approximate multipliers} to MACs in SA tiles $T=\{T_1,T_2,...\}$, i.e., mapping $map_{TM}: T \rightarrow M$, and for the \textit{assignment of the convolutional layers} to SA tiles, i.e., mapping $map_{LT}: L \rightarrow T$. The weights in each layer are updated according to the properties of a particular multiplier assigned to the tile which computes the output of the layer.


\begin{figure}[ht]
    \centering\vspace{-1em}
    \includegraphics[width=0.9\columnwidth]{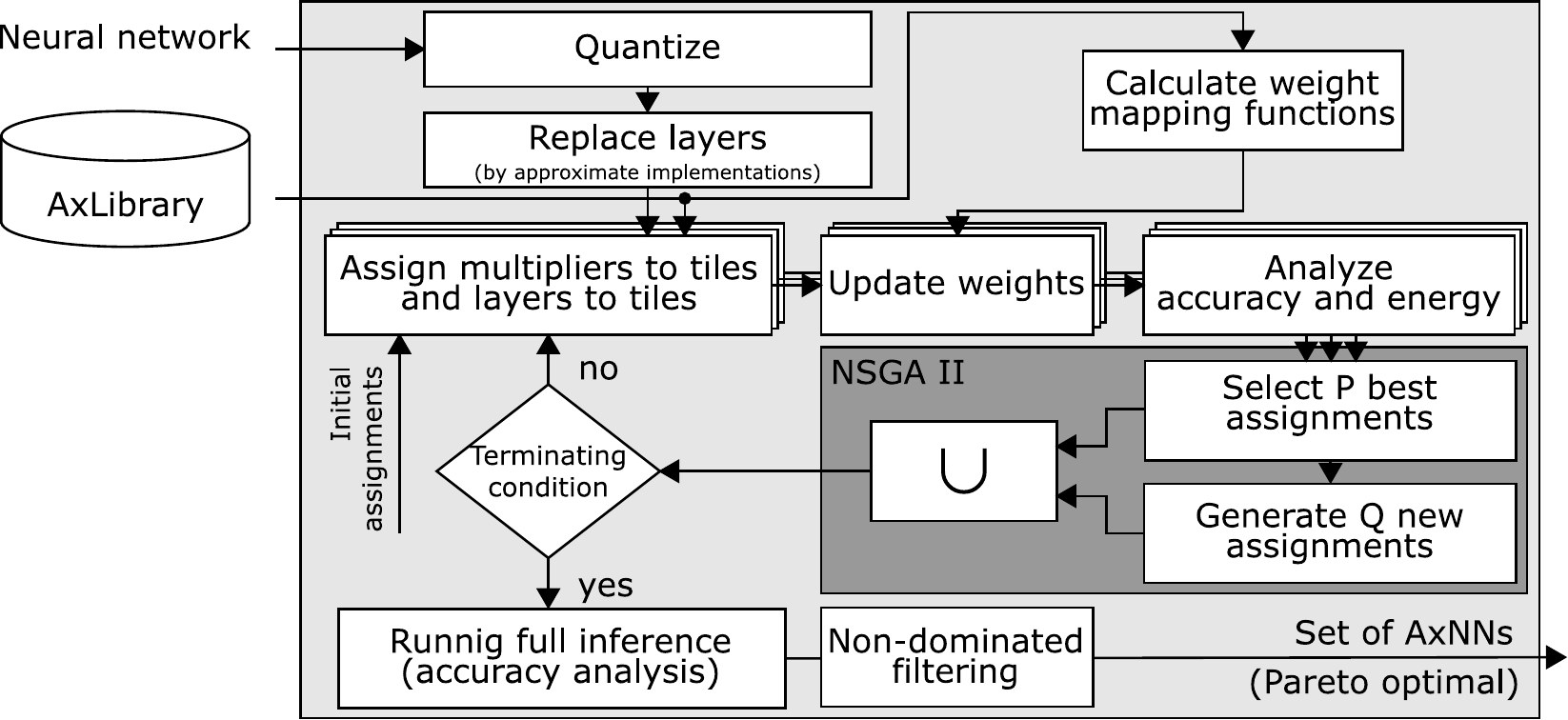}\vspace{-1em}
    \caption{Overall architecture of \toolname framework }\vspace{-1.5em}
    \label{fig:overall}
\end{figure}

The overall architecture of the proposed framework is shown in Figure \ref{fig:overall}.
The framework expects that a fully specified NN is available (typically in \textit{protobuf} format). If not already done, the NN is firstly quantized to avoid floating point MAC operations. 
The protobuf specification of the quantized NN is then edited and all convolutional layers are replaced by approximate ones. This step is necessary to have the ability to specify which multiplier should be used to calculate the output of the MACs separately for each layer. 
To obtain a Pareto set of various AxNNs, we propose to use multi-objective genetic algorithm (NSGA-II)~\cite{deb:2002}. The algorithm maintains a population of $|P|$ candidate solutions represented as a pair $(map_{TM},map_{LT})$. 
The search starts from an initial population which is generated either deterministically or randomly. 
The candidate solutions are iteratively optimized with respect to the accuracy of AxNN and energy required to perform one inference. 
For each candidate solution, a corresponding protobuf is created. This step includes the assignments of the multipliers to each approximate layer according to the $map_{TM}$ and $map_{LT}$ and refinements of the weights in each approximate layer depending on the chosen multiplier. Then, energy as well as quality of the obtained AxNN is evaluated on a subset of training data. 
The usage of the subset of training data reduces the testing time and it simultaneously avoids over-fitting. 
At the end of the optimization process when a terminating condition is met (typically the maximum number of allowed iterations is exceeded), the quality of the candidate solutions is evaluated using the complete training set. Solutions whose parameters are dominated by at least one other solution are filtered out.



\subsection{Weight tuning}
Weight tuning represents one of the essential steps and key advantage of the proposed framework. Updating the weights of approximate NN helps us to avoid a time consuming retraining which is typically inevitable to restore the accuracy of NNs after introducing approximate components~\cite{Venkataramani:axnn}.
The objective of the weight tuning is to replace weights with different values layer by layer in such cases where the new weights will potentially lead to better overall accuracy.
As this problem itself is nontrivial, the potential candidates for replacement are determined according to the properties of the multipliers assigned to each layer independently on the structure of the NN or data distribution. 
We exploit the fact that the value of the second operand is constant for each multiplication while the value of the first operand differs depending on the input data.
It is hypothesized that the lower arithmetic error of the multiplier for a particular weight leads to higher accuracy of the whole AxNN.

Weight tuning is based on the knowledge of a \textit{weight mapping function} $map_{M_i}$ precalculated offline for each approximate multiplier $M_i \in M$.
Let $W \subset \mathbb{N}$ be the range of weight values and $I \subset \mathbb{N}$ be the range of data values (typically $W = I = \{0,...,2^k-1\}$ for a $k$-bit quantized NN), $map_{M_i}$ is then determined as
$$\forall w \in W: map_{M_i}(w) =  \argmin_{w' \in W} \sum_{a \in I}{|M_i(a,w') - a\cdot w|}.  $$
It means that for each weight $w$, a weight $w'$ is determined that minimizes the sum of absolute differences between the output of the approximate and accurate multiplication over all inputs $a\in I$. As the size of $I$ is constant, \textit{mean error distance} (MED) is in fact minimized.
If $M_i$ is an accurate multiplier, then the equation implies that  $w' = w$.

The update weight procedure works as follows.
Each weight $w$ in layer $l \in L$ is replaced by the output of $map_{M_i}(w)$, where $M_i = map_{TM}( map_{LT}(l) )$. Since $|I|=2^k$, this approach is applicable approximately to $k \leq 12$ because of memory requirements.

\subsection{Representation of candidate AxNNs}
\begin{figure}[ht]
    \centering
    \includegraphics[width=0.9\columnwidth]{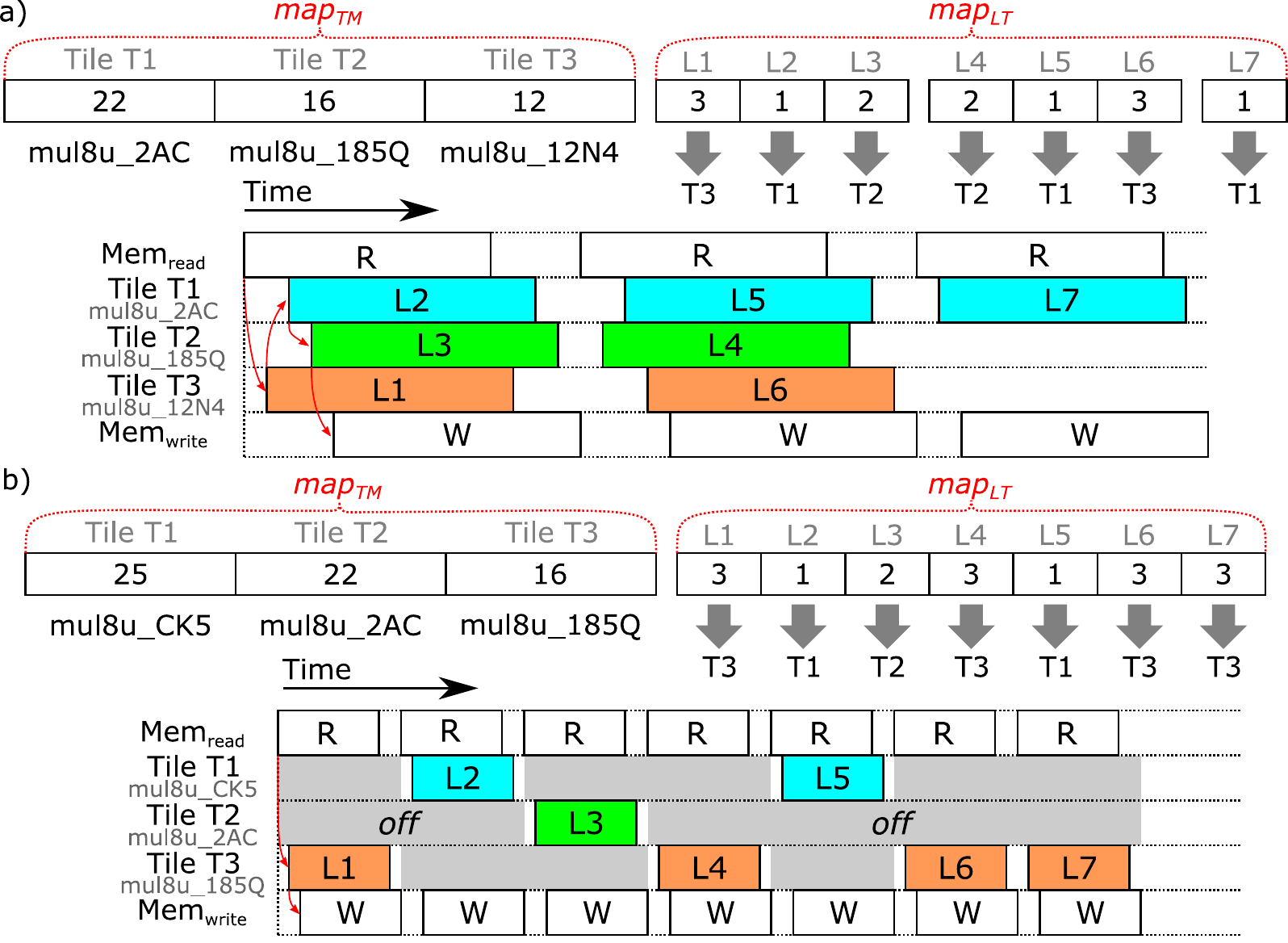}\vspace{-1em}
    \caption{Example of two encodings of different AxNNs for NN with $|L|=7$ approximate convolution layers for (a) pipelined and (b) power-gated HW accelerators, both with $|T|=3$ tiles. For each AxNN, the corresponding execution plan is shown. For better clarity, only the computation of the convolution layers is depicted. The data dependencies are shown for the first step using red arrows. Grey area in (b) indicates that the corresponding tile is switched off.
    The white area indicates no workload due to pipeline dependencies (a) or time required to switch the tile on (b). As shown, all tiles are always (except of the last step) active in (a). In (b) the architecture allows to reuse the same tile more than once (see the allocation of L6 and L7).
    }
    \label{fig:representation}\vspace{-1em}
\end{figure}

Each candidate solution is uniquely defined by a pair $(map_{TM},map_{LT})$.
We propose to use an integer-based encoding. 
The first part, $map_{TM}$ is encoded using $|T|$ integers where each integer corresponds with index $i$ of multiplier $M_i \in M$. Similarly, the second part is encoded using $|L|$ integers where each integer determines index $i$ of a tile $T_i \in T$ that will be used to compute the output of the corresponding layer.
Depending on the structure of the chosen HW accelerator, additional restrictions may be applied.
For the power-gated architecture there are no additional requirements either on $map_{TM}$ or $map_{LT}$ because only one tile is active at any moment. The remaining tiles are suspended. On the other hand, all $|T|$ tiles are requested to have a workload in the pipelined architecture. 
This requirement puts a constraint on $map_{LT}$. If we divide the encoding into chunks consisting of $|T|$ integers, each chunk must encode a permutation of the set $\{1,2,\dots,|T|\}$.
An example of AxNN encoding for both architectures is given in Figure~\ref{fig:representation}.

\subsection{Design space exploration}
All AxNNs that are generated by $(map_{TM},map_{LT})$ pairs give arise a large search space which is unfeasible to be enumerated exhaustively. 
Suppose we have a typical NN which consists of 50 layers and a library of 20 approximate multipliers. Let $|T|=4$, for example. Then there exists more than $1.2\cdot10^{22}$ different AxNNs for pipelined architecture and more than $2.0\cdot10^{35}$ different AxNNs for power-gated architecture.
Hence, we propose to employ a heuristic algorithm for the design space exploration. 
Because we are looking for AxNNs optimized for two criteria (energy and accuracy), the multi-objective genetic algorithm is naturally a method of the first choice. 

\begin{figure}[ht]
    \centering\vspace{-1em}
    \includegraphics[width=0.8\columnwidth]{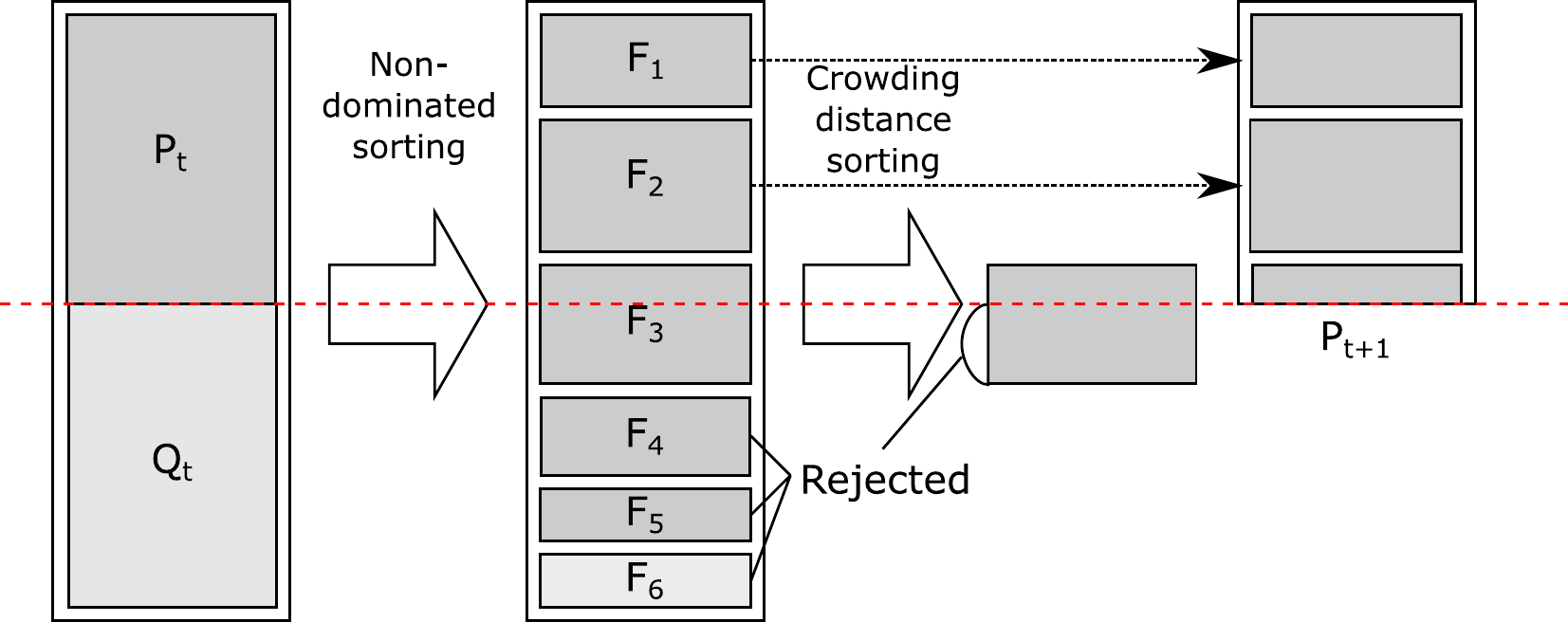}\vspace{-1em}
    \caption{Creating of a new population in NSGA-II algorithm~\cite{deb:2002}}
    \label{fig:nsga}
\end{figure}

We are primarily interested in AxNNs belonging to a Pareto set which contains the so-called \textit{nondominated solutions}. The dominance relation is defined as follows. Consider two AxNNs $N_1$ and $N_2$. Network $N_1$ dominates another network $N_2$ if: (1) $N_1$ is no worse than $N_2$ in all objectives, and (2) $N_1$ is strictly better than $N_2$ in at least one objective.
According to the literature, the most powerful optimization method for a small number of design objectives is a Non-dominated Sorting Genetic Algorithm (NSGA-II) \cite{deb:2002}.
In each iteration, NSGA-II maintains a population of candidate solutions $P_t$ of fixed size. The current population $P_t$ is used to generate a set of offspring $Q_{t}$ that are subsequently evaluated. All $P_t \cup Q_t$ individuals are then sorted according to the dominance relation into multiple fronts (see Fig. \ref{fig:nsga}). 
The first front $F_1$ contains all non-dominated solutions along the Pareto front. Each subsequent front ($F_2$, $F_3$, \dots) is constructed similarly but from individuals that are not included in the previous fronts. The first fronts ($F_1$ and $F_2$ in Fig. \ref{fig:nsga}) are copied to the next population $P_{t+1}$. If any front must be split ($F_3$ in Fig. \ref{fig:nsga}), a crowding distance (please see \cite{deb:2002} for details) is used for the selection of individuals that are copied to $P_{t+1}$.
The key advantage of this algorithm is that it re-constructs the Pareto front in each iteration and tries to cover all possible compromise solutions.

To generate $Q_t$ offspring from $P_t$ candidate solutions, we propose to use \textit{uniform crossover} operator followed by \textit{mutation} operator. The crossover randomly picks two individuals (so called parents) from $P_t$ and produces a single candidate solution which combines information from both parents. In our case, a new string of integers is produced where each integer is chosen from either parent with equal probability. Then, with small probability $p_{mut}$ one integer in the obtained candidate solution is randomly changed respecting the constraints.


\section{Experimental setup}\label{sec:exsetup}
\begin{figure}[b]
    \centering\vspace{-1em}
    \includegraphics[width=\columnwidth]{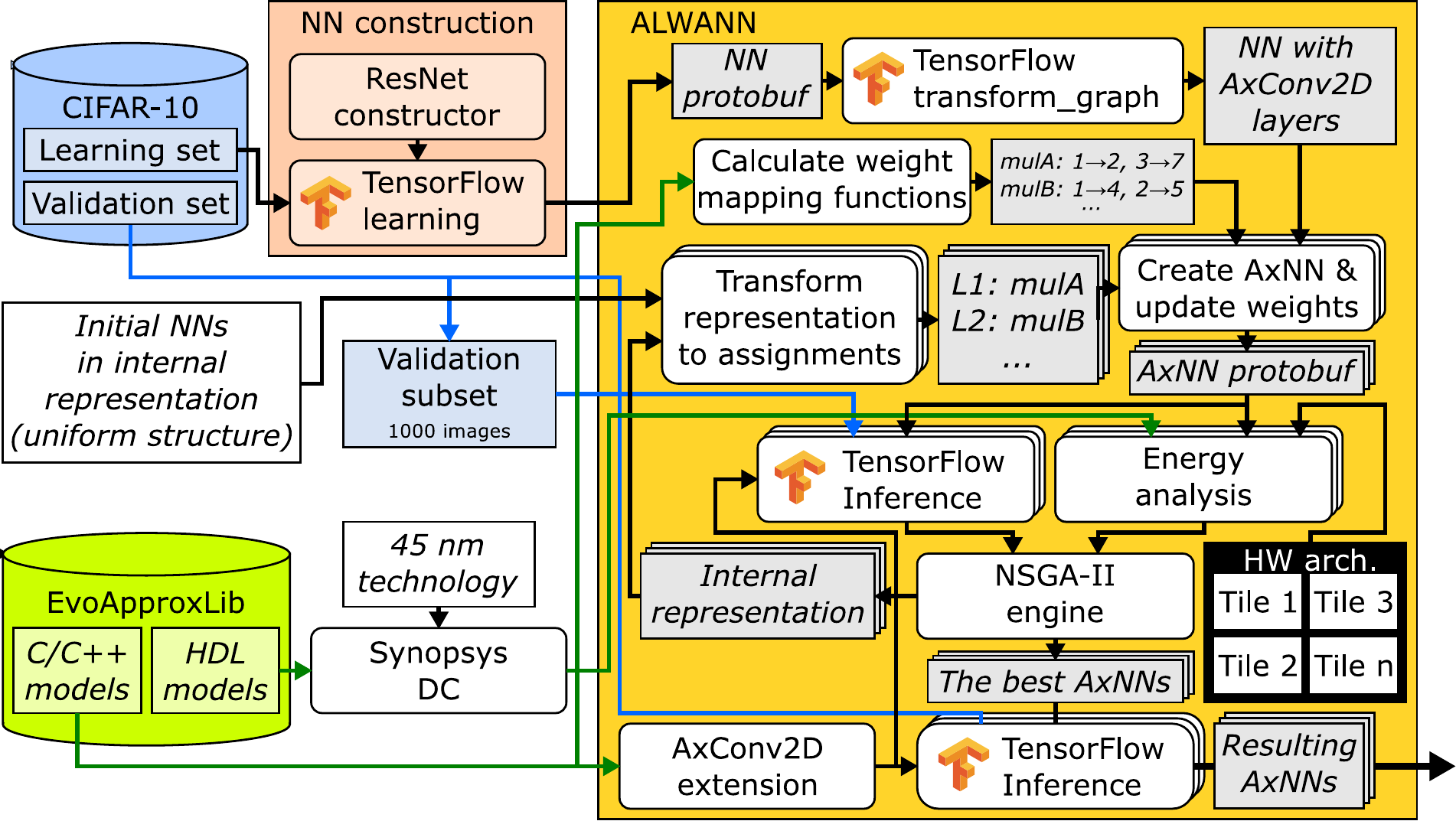}\vspace{-1em}
    \caption{Our toolflow for retraining-less approximation of ResNet neural network.}\label{fig:toolflow}
    \vspace{-1em}
\end{figure}

To evaluate \toolname, we extended TensorFlow framework to support approximate quantized layers. The tool flow is shown in Figure~\ref{fig:toolflow}. At the beginning, the common \texttt{QuantizedConv2D} layers are replaced with newly introduced \texttt{AxConv2D} layers. The remaining part follows the scheme already described in Section~\ref{sec:methodology}. For the evaluation, ResNet networks (v1 with non-bottleneck blocks)~\cite{he:resnet} were chosen and trained to recognize images from CIFAR-10 dataset. The reasons are threefold -- we want to demonstrate the scalability of \toolname (ResNet consists of many layers and multipliers), evaluate the quality drop of the obtained AxNNs (ResNet offers a reasonable classification accuracy) and finally, compare the parameters of AxNNs with various ResNet instances because this NN can be naturally approximated by reducing the number of convolution layers. 
Three ResNet instances with different number of layers were constructed and trained in $10^5$ epochs. The resulting NNs were frozen, quantized and convolutional layers were replaced by approximate multipliers by means of \textit{transform\_graph} tool. Table~\ref{tab:nn} shows the number of convolutional layers, the number of multiplications executed in these layers for inference of one input image ($32\times32$ RGB pixels), the classification accuracy of the floating-point NN and the accuracy after quantization.


\begin{table}[ht]
    \centering\vspace{-1em}
    \caption{Parameters of ResNet NNs considered in experiments}
    \label{tab:nn}\vspace{-1em}
    \begin{tabular}{l c c c c}\toprule
         \bf ResNet & \bf \# conv. & \bf \# & \bf accuracy & \bf accuracy   \\
         \bf instance  & \bf layers & \bf mults. & \bf (floating-point) & \bf (qint-8) \\\midrule
         ResNet-8 &  7 &  21.1M & 83.42\% & 83.26\% \\
         ResNet-14 &  13 &  35.3M & 85.93\% & 85.55\% \\
         ResNet-50 &  49 &  120.3M & 89.38\% & 89.15\% \\\bottomrule
    \end{tabular}
\end{table}

The library of approximate multipliers consists of all 36 eight-bit multipliers from the publicly available EvoApproxLib library~\cite{mrazek:date16lib}.
The energy of each multiplier was computed using Synopsys DC and $45~nm$ fabrication technology with uniform input distribution. 

The search is initialized with the population of $|P_0|=36$ different AxNNs with \textit{uniform architecture} (each AxNN uses exactly one of the 36 multipliers in all layers). 
In each iteration, $|P_i| = 50$ best solutions are chosen and $|Q_i| = 50$ new candidate AxNNs are generated. The probability of mutation is $p_{mut} = 10\%$. The experiments are run separately for power-gated and pipelined HW architecture having $2 \leq |T| \leq 4$ tiles for AxResNet-8 (approximate implementation of ResNet-8) and AxResNet-14, and $3 \leq |T| \leq 6$ tiles for AxResNet-50. The energy of candidate AxNNs was estimated as the number of multiplications in the layer multiplied by the average energy of multiplication in the layer. 
First $1,000$ out of $10,000$ validation images were used by the search algorithm which was executed for $30$ iterations. Finally, the inference was executed for the full validation set. 
All the experiments were performed on Intel E5-2630 CPU running at 2.20 GHz. 
Training of the initial ResNet neural network took 48 hours in the average. 

\section{Results}\label{sec:results}
\subsection{Impact of weight tuning}

\begin{figure}[ht]
    \centering\vspace{-1em}
    \includegraphics[width=\columnwidth]{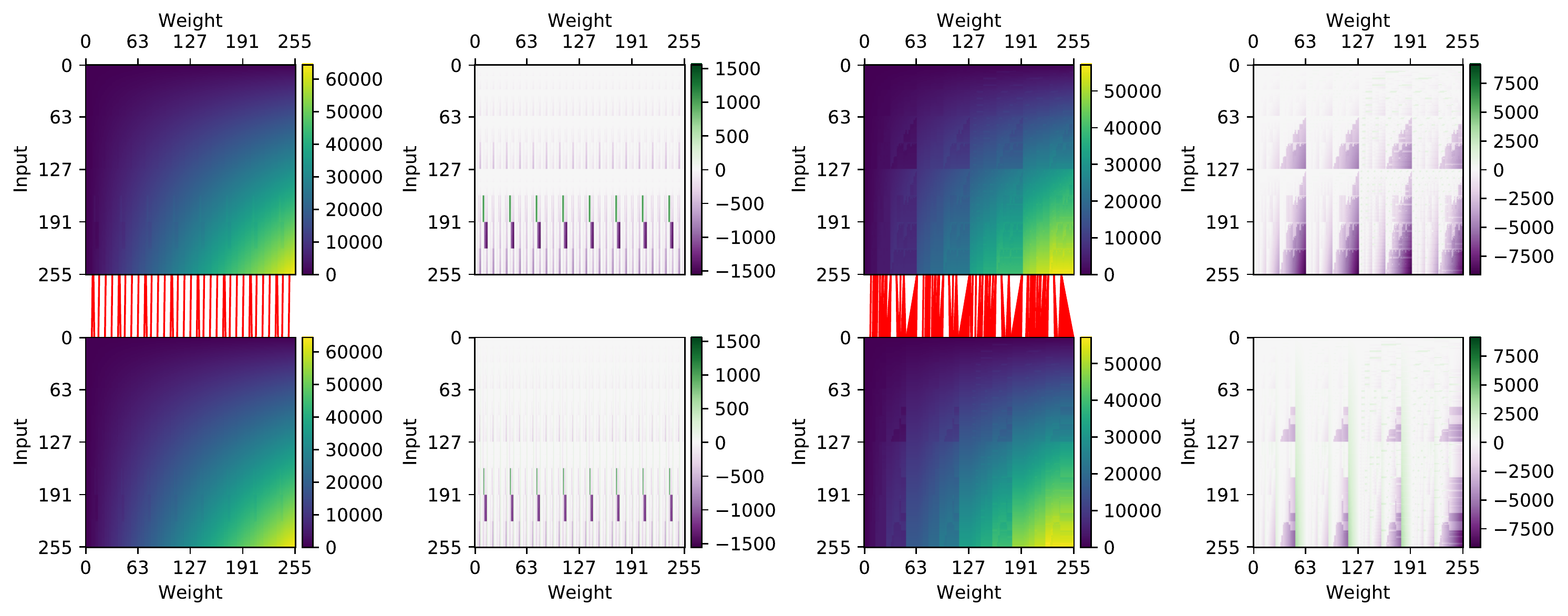}\vspace{-1em}
    \caption{Output values \textit{(1\textsuperscript{st} and 3\textsuperscript{rd} columns)} and differences to accurate multiplier \textit{(2\textsuperscript{nd} and 4\textsuperscript{th} columns)} of approximate multiplier \texttt{mul8u\_7C1} (first two columns) and \texttt{mul8u\_L40} before \textit{(upper row)} and after \textit{(lower row)} applying weight-mapping to the weight. The red lines show the mapping $map_{\mathrm{mul8u\_7C1}} = \{0 \rightarrow 0; \dots; 7 \rightarrow 8; 10 \rightarrow 9; \dots; 247 \rightarrow 248\}$ and $map_{\mathrm{mul8u\_L40}}=\{0 \rightarrow 0; \dots; 7 \rightarrow 8; 10 \rightarrow 11; \dots; 237\mathrm{-}255 \rightarrow 240 \}$. }
    \label{fig:mul8u}
\end{figure}

First, we analyzed how MED of approximate multipliers and accuracy of AxResNet-8 are influenced when we apply the proposed weight mapping method on the value fed to the second input of the multiplier. 
Figure~\ref{fig:mul8u} illustrates how the output values and errors are changed (with respect to the exact multiplication) after applying the mapping for two chosen approx. multipliers. For mul8u\_7C1, $39$ weights are changed by $\pm 1$ which reduced MED from $87.3$ to $69.7$ (-20\%). The MED of the multiplier mul8u\_L40 was improved from $1011.3$ to $647.7$ (-36\%). 
The computation of the optimal mapping $map_M$ takes around 0.10 seconds in the average. 

When the mapping is applied to various manually (algorithmically) created configurations of AxResNet-8 (one layer approximated and the rest remains accurate; one layer is accurate and the rest is approximated; all layers are approximated; the same approx. mult. used in all approx. layers), the average improvement of the classification accuracy of 370 out of 510 AxNNs was 8.2\%. In the remaining cases, the accuracy dropped by 1.4\% in the average. Considering all AxNNs, the average change of the accuracy was +5.0\%.

\begin{figure}[ht]
    \centering\vspace{-1em}
    \includegraphics[width=\columnwidth]{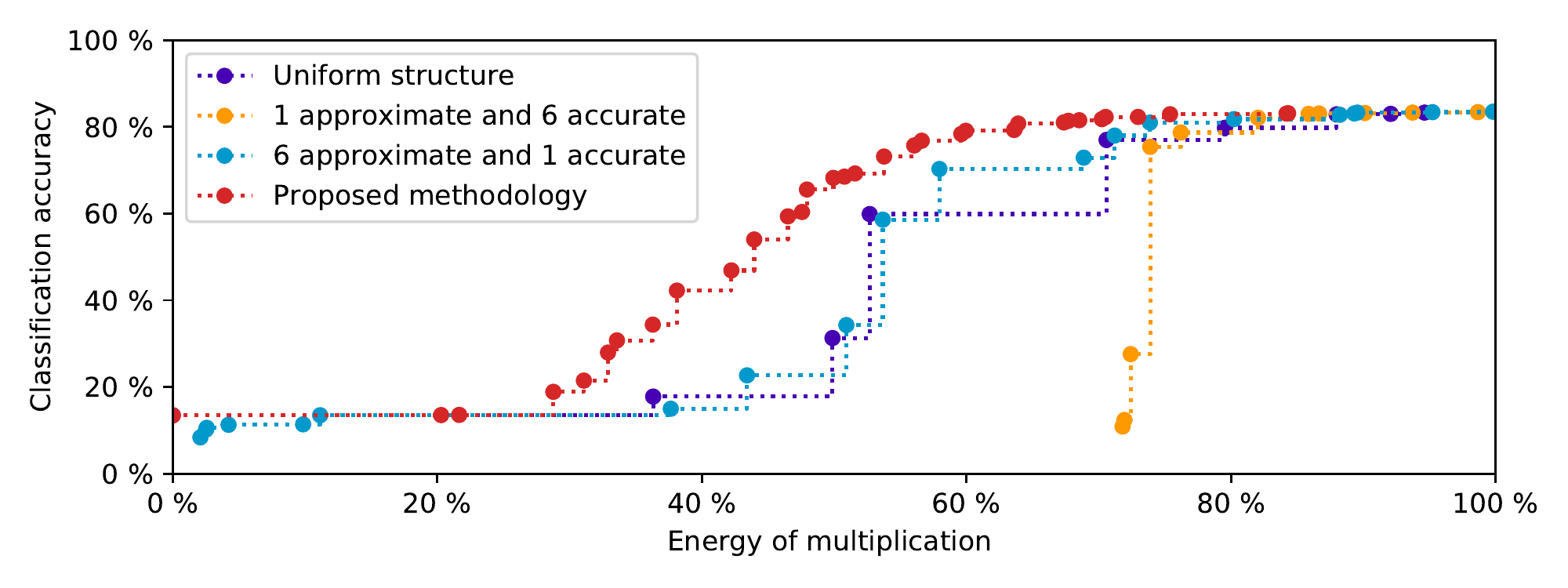}\vspace{-1em}
    \caption{Comparison of AxResNet-8 approximate neural networks constructed by means of proposed algorithm and NNs having a regular structure.}
    \label{fig:r8}\vspace{-1.5em}
\end{figure} 

\subsection{Quality of the generated AxNNs}
Figure~\ref{fig:r8} shows the quality of AxNNs obtained using \toolname from the original ResNet-8. The results are compared with three configurations of AxNNs mentioned in the previous section. 
The proposed method delivers significantly better AxNNs compared to the manually created AxNNs.
The uniform structure (all layers approximated) widely used in the literature (see e.g.,~\cite{Sarwar:2018,mrazek:iccad16}) achieves results comparable to AxNNs with all but one approximated layers. In contrast to that, AxNN with one approximate layer leads to significantly worse results because of small energy saving.
The proposed method provides \textbf{better trade-offs between the accuracy and energy consumption} in comparison with the uniform NN architectures reported in the state-of-the-art works.

\begin{figure}[ht]
    \centering\vspace{-1em}
    \includegraphics[width=0.95\columnwidth]{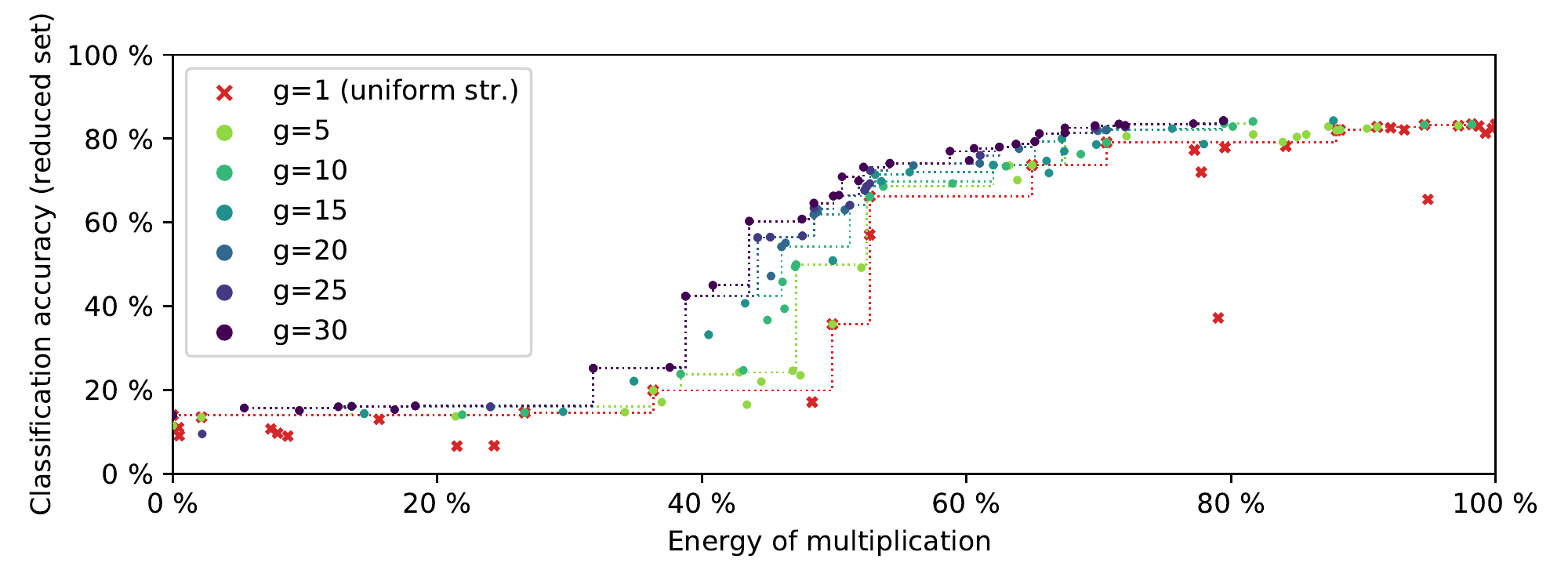}\vspace{-1em}
    \caption{Evolution of candidate AxNNs (ResNet-8, pipelined arch., 4 tiles) over generations (g). Red crosses show the initial (uniform) AxNNs.\vspace*{-1.5em}}
    \label{fig:gen}
\end{figure}

\begin{figure*}
    \centering\vspace{-1em}
    \includegraphics[width=\textwidth]{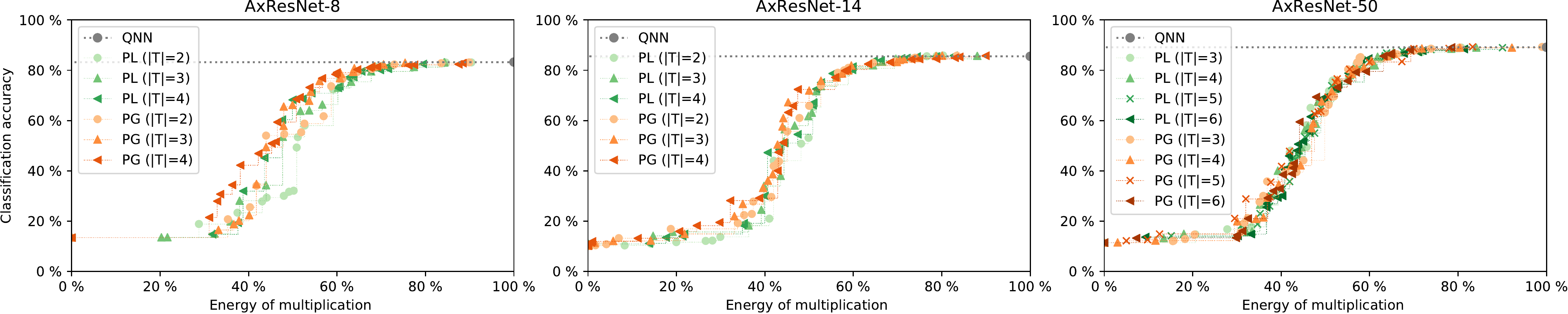}\vspace{-1em}
    \caption{Accuracy and energy of the best AxNNs constructed for both acceleratos architectures --- pipelined (PL) and power-gated (PG) with various tiles count ($|T|$). The AxNNs are compared with the original accurate, but quantized NN (QNN).}
    \label{fig:all}\vspace{-1.5em}
\end{figure*}

Figure~\ref{fig:all} illustrates the impact of the architectural constraints (pipelined or power-gated, the number of tiles) on the quality of resulting AxNNs. In contrast to pipelined architectures, the power-gated architectures allow us to construct more energy-efficient AxNNs because there is no constraint on the tiles load balance. For instance, there are $10^3 \times$ more possible options for PG AxResNet-14 ($10^{11} \times$ for AxResNet-51) with $|T|=3$ than for PL architectures. Similarly, increasing the number of tiles leads to more options in selecting the approximate multipliers and thus more effective solutions.
Figure~\ref{fig:gen} shows the evolution of candidate solutions for one selected hardware configuration. The search converges towards better solutions and the most important changes are introduced in first iterations of the search algorithm.

We also analyzed the multipliers employed in the resulting AxNNs. We considered NNs having the classification accuracy $>80\%$ only. In AxResNet-50 and AxResNet-14 NNs, four approximate multipliers {\footnotesize(mul8u\_19DB, mul8u\_NGR, mul8u\_2HH, mul8u\_QJD)} have appeared in 78\% and 71\% layers respectively. In AxResNet-8 more different approximate multipliers have occurred; four the most frequent multipliers {\footnotesize (mul8u\_185Q, mul8u\_2AC, mul8u\_CK5, mul8u\_GS2)} are employed in 64\% of layers. These multipliers consume 52--90\% of energy, MED after tuning is 0.015--0.9\%, the error probability is higher than 90\% (except one multiplier in each set) and the mean relative error varies from 0.5 to 4.5\%. It is remarkable that the worst-case error distance is 0.05--1.2\% (0.5\% in the average) for the smallest network, but the larger network only shows half error: 0.06--0.6\% (0.29\% in the average). 

A bottleneck of the algorithm is the expensive simulation of approximate multipliers on CPU. Although the multipliers were cached, our single core application has 10x lower performance than vectorized accurate multiplication. Table~\ref{tab:speed} gives the time spent on a single CPU to evaluate (i.e., to run the inference) of one and all AxNNs in the search and validation parts and the total time to obtain one Pareto front of AxNNs for the given configuration. Note that we run multiple configurations on one multi-core CPU. Almost the whole execution time is spent in the inference part; requirements of the other procedures of the tool flow are negligible. The slow inference can effectively be solved by using an approximate reconfigurable accelerator. If we assume that such an accelerator can accelerate the inference more than 1400x (4x faster than GPU~\cite{Nurvitadhi:2017} which has 35x better performance that our CPU employed in the experiments where we utilized only one core out of 10), the approximation of ResNet-50 would take 10 minutes only.

\begin{table} [ht]
    \centering\vspace{-0.5em}
    \caption{Time requirement for \toolname algorithm}
    \label{tab:speed}\vspace{-1em}
    \begin{tabular}{c c c c c c}\toprule
        \multirow{2}{*}{\bf AxNN} & \multicolumn{2}{c}{\bf Searching} & \multicolumn{2}{c}{\bf Validation} & \multirow{2}{*}{\bf Total} \\\cmidrule(lr){2-3}\cmidrule(lr){4-5}
         & one eval. & total &  one eval. & total & \\\midrule
         AxResNet-8    & 25 sec    & 10.4 h    & 7.4 min  & 6.2 h   & 0.7 days \\
         AxResNet-14   & 100 sec   & 41.7 h    & 14.5 min & 12.1 h  & 2.3 days \\
         AxResNet-50   & 322 sec   & 134.2 h   & 54.5 min & 45.4 h  & 7.5 days \\\bottomrule
    \end{tabular}\vspace{-1em}
\end{table}

%


\subsection{Overall results}

\begin{table}[ht]\setlength{\tabcolsep}{2pt}
    \centering
    \caption{Parameters of selected AxNNs} 
    \vspace{-1em}
    \label{tab:res} 
    \begin{tabular}{l r r r r r} 
        \toprule
        \bf AxNN & \bf Accuracy & \bf Relative accuracy & \bf Relative energy & \bf Total energy \\\midrule
         \parbox[t]{2mm}{\multirow{7}{*}{\rotatebox[origin=c]{90}{AxResNet-50}}} 
         &\it   89.15 \% &\it 100.00 \% &\it 100.00 \%   &\it 120.27 M \\
         &   89.30 \% & 100.17 \% & 83.29 \%   & 100.17 M \\
         &   89.08 \% & 99.92 \% & 78.47 \%   & 94.37 M \\
         &   88.69 \% & 99.48 \% & 77.97 \%   & 93.77 M \\
         &   88.58 \% & 99.36 \% & 70.02 \%   & 84.21 M \\
         &   88.10 \% & 98.82 \% & 69.12 \%   & 83.13 M \\
         &    87.77 \% & 98.45 \% & 67.36 \%   & 81.02 M \\
         &   85.00 \% & 95.34 \% & 57.74 \%   & 69.45 M \\

        \midrule
           
        \parbox[t]{2mm}{\multirow{5}{*}{\rotatebox[origin=c]{90}{AxResNet-14}}} &\it   85.55 \% &\it 100.00 \% &\it  100.00 \% &\it 35.33 M \\
        &      85.87 \% & 100.37 \% &  80.32  \% & 28.38 M \\
        &      85.42 \% & 99.85 \% &   74.34  \% & 26.27 M \\
        &      84.77 \% & 99.09 \% &   70.85  \% & 25.04 M \\
        &      83.82 \% & 97.98 \% &   64.64  \% & 22.84 M \\
           \midrule

        \parbox[t]{2mm}{\multirow{5}{*}{\rotatebox[origin=c]{90}{AxResNet-8}}} &\it 83.26 \% &\it  100.00 \% &\it 100.00 \% &\it 21.18 M \\
        & 83.16 \% &   99.88 \% & 84.31  \% & 17.86 M \\
        & 81.79 \% &   98.23 \% & 70.23  \% & 14.87 M \\
        & 79.11 \% &   95.02 \% & 59.95  \% & 12.70 M \\
        & 75.71 \% &   90.93 \% & 56.04  \% & 11.87 M \\
        \bottomrule
    \end{tabular}\vspace{-1em}
\end{table}

Table~\ref{tab:res} gives some parameters of the best AxNNs constructed using the proposed tool. The following parameters are reported for each network: relative accuracy, total and relative energy of convolutional operations. The relative values are calculated with respect to the original quantized (8-bit) ResNet.
The quality of the obtained AxNNs for ResNet-50 is very promising. If a target application is able to tolerate 1\% accuracy drop (from 89.15\% to 88.1\%), for example, we can save more than 30\% of energy.
The evaluation across different architectures shows that it is not advantageous to use AxNNs having more than 4\% (2\% for AxResNet-14) degradation of accuracy for AxResNet-50, because AxResNet-14 (AxResNet-8) exhibit the same quality but lower energy. 

\begin{figure}[b]
    \centering\vspace{-1.8em}
    \includegraphics[width=\columnwidth]{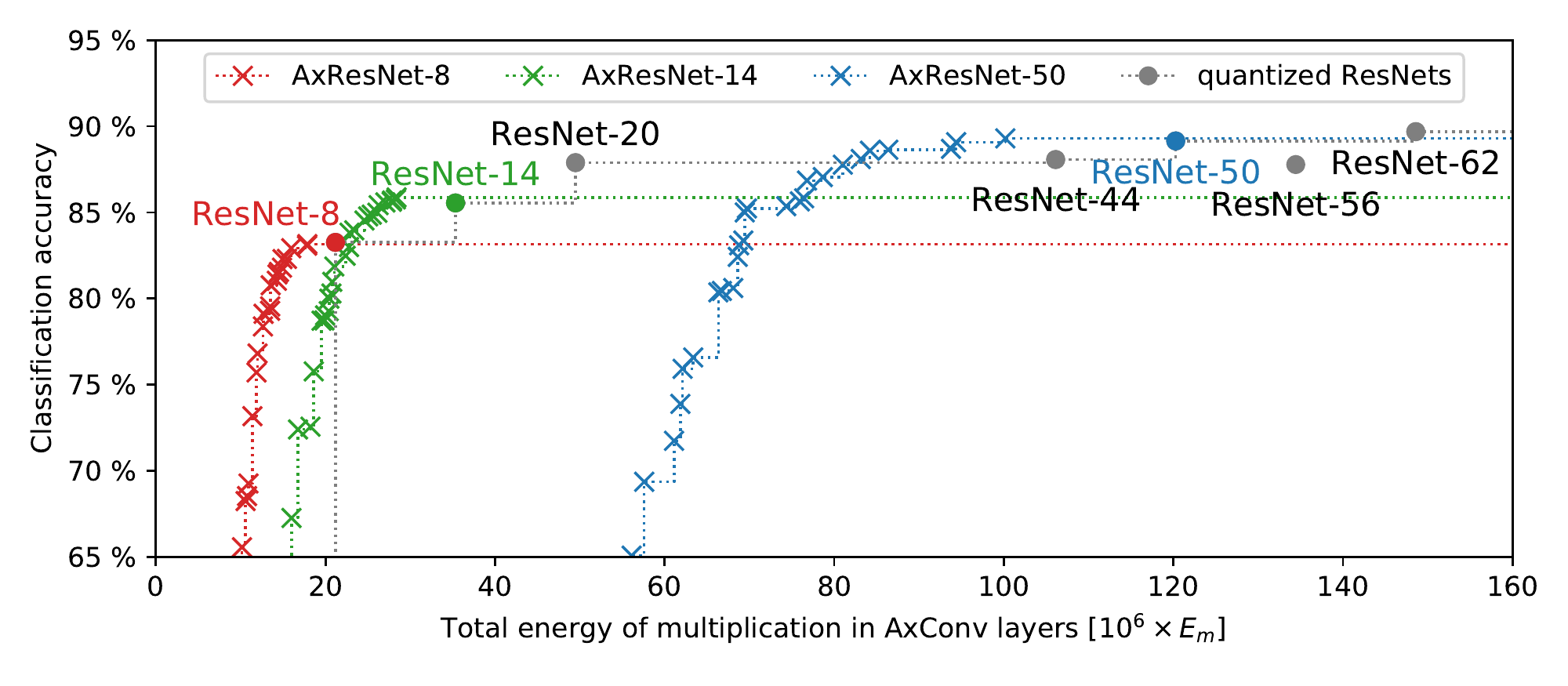}\vspace{-1em}
    \caption{Comparison of proposed AxNNs (crosses) with accurate quantized NNs (points) --- the energy reports the energy of multiplications in the convolutional layers while $E_{m}$ is energy of one multiplication. Gray points represent quantized networks that were not approximated (complexity reduction).}
    \label{fig:allone}
\end{figure}

Complete overview of the the best obtained AxNNs having accuracy higher than 65\% is provided in 
Figure~\ref{fig:allone}. 
In addition to the parameters of the AxNNs for three ResNet architectures discussed so far, we included also the parameters of all possible ResNet architectures up to 62 layers (see the dots), namely ResNet-20, -44, -56 and -62, that have been trained in the same way as the remaining ResNet NNs. 
These NNs have been obtained by reducing the number of layers by multiples of six, i.e., at block boundaries.
In total, 7 different ResNet architectures are included.
As evident, our method is able to produce significantly more design points; more than 40 points are produced from a single ResNet.
Moreover, majority of the design points are unreachable by simple reduction of the number of layers (see the blue crosses vs. dot symbols).
Considering the computational complexity, each ResNet instance must be trained separately. 
For complex structures, training of a new structure can took several days or weeks on computer clusters. 

\subsection{Comparison with SoA}

Table~\ref{tab:compare} compares the proposed approach with the state of the art approaches for reducing the energy of NNs that have been evaluated on CIFAR-10 dataset. 
Table~\ref{tab:compare} includes reported energy reduction and accuracy degradation. The requirement for retraining, uniformity of the architecture and complexity of NN are also provided. In contrast with multiplier-less multiplication where only 4 different architectures were proposed~\cite{Sarwar:2018}, our approach allows to find a new design points with high granularity without retraining. Besides that, our approach enabled us to find AxNNs with low energy exhibiting low accuracy, e.g., $<$80\%. Even these solutions can be beneficial, for example as one of initial stages of some Progressive Chain Classifier~\cite{Choi2019}.

In~\cite{mrazek:iccad16}, where arbitrary approximate multiplier could be employed, there were 10 retraining steps taken to improve quality of LeNet-6 ($\sim$278k mults.). Since ResNet-8 network is 75x larger, we can assume, that evaluation of one AxNN would take 15 days instead of 25 seconds needed in \toolname. The ResNet-50 network (431x larger) would be evaluated in 89 days instead of 322 seconds. During the search, we had to evaluate 1,500 candidate solutions. It is clear, that it is \textbf{unfeasible to perform retraining} for large AxNNs employing   \textbf{arbitrary approximate multipliers}. 

\begin{table}\setlength{\tabcolsep}{2pt}
    \centering
    \caption{Comparison of automated NN approximation methods: architectural parameters, energy and accuracy reduction reported on CIFAR-10}\vspace{-1em}
    \label{tab:compare}
    \footnotesize
    \renewcommand*{\arraystretch}{1.1}%
    \begin{tabular}{l c l}\toprule
\bf Approach & \bf Retrain. / Unif. / Depth & \bf Energy / Accuracy  \\\midrule 
\multirow{2}{*}{Venkataramani~\cite{Venkataramani:axnn}} & \multirow{2}{*}{yes / no / low} & -22\% / -0.5\% \\
 &  & -26\% / -2.5\% \\\midrule
Sarwar~\cite{Sarwar:2018} & yes / yes / high & -33\% / -1.8\% \\\midrule
 &  & -12\% / -1.2\% $_{50 \rightarrow 44}$ \\
He~\cite{he:resnet}      &  yes / yes / high  & -71\% / -4.0\%   $_{50 \rightarrow 14}$ \\
       &                & -48\% / -2.7\%   $_{14 \rightarrow 8}$ \\\midrule
       
  &  & -30\% / -0.6\% $_{\text{AxRN-50}}$ \\
  This paper & no / no / high  &            -30\% / -0.9\%  $_{\text{AxRN-14}}$ \\
&     &            -30\% / -1.7\%  $_{\text{AxRN-8}}$ \\\bottomrule
    \end{tabular}\vspace{-1.5em}
\end{table}

\section{Conclusion}
The proposed methodology \toolname allows us to approximate hardware accelerators of convolutional neural networks and thus optimize the energy consumption of the inference path of DNNs. We achieved better energy savings with the same accuracy as the other algorithms that employ retraining. The retraining typically results in (i) approximation of significantly smaller networks (limited scalability)~\cite{mrazek:iccad16,zhang:2015}, or (ii) limited set of considered approximate components~\cite{Sarwar:2018}. 
The proposed fast weight-mapping algorithm allows us to adapt the network to the approximation errors without any processing of the input data. 
The proposed methodology can enable us in the future to improve the energy efficiency of DNN hardware in real-time.

\vspace{1mm}
{\smaller
\noindent\textit{Acknowledgment}
This work was supported by Czech Science Foundation project 19-10137S and by the Ministry of Education of Youth and Physical Training from the Operational Program Research, Development and Education project International Researcher Mobility of the Brno University
of Technology --- CZ.02.2.69/0.0/0.0/16\_027/0008371\par
}
\vspace{-1mm}

\bibliographystyle{IEEEtran}
\bibliography{IEEEabrv,iccad19} 

\end{document}